\pdfoutput=1

\documentclass[11pt]{article}

\usepackage[preprint]{acl}

\usepackage{times}
\usepackage{latexsym}

\usepackage[T1]{fontenc}

\usepackage{xspace}

\usepackage[utf8]{inputenc}

\usepackage{microtype}

\usepackage{inconsolata}

\usepackage{graphicx}

\title{Conversational Prompt Engineering}

\author {
Liat Ein-Dor\thanks{\ \ These authors equally contributed to this work.}, Orith Toledo-Ronen\footnotemark[1], Artem Spector\footnotemark[1],
Shai Gretz\footnotemark[1],
Lena Dankin\footnotemark[1],\\ {\bf Alon Halfon,} {\bf Yoav Katz,} {\bf Noam Slonim}\\\\
 IBM Research \\
 \
 \{liate, oritht, artems, avishaig, lenad, alonhal, katz, noams\}@il.ibm.com}

\usepackage{algorithm}
\usepackage{algpseudocode}

\begin{document}

\maketitle
\newcommand{\userComment}[2]{{\color{blue} \textbf{#1}: #2}}
\newcommand{\led}[1]{\userComment{Liat}{#1}}
\newcommand{\ot}[1]{\userComment{Orith}{#1}}
\newcommand{\sg}[1]{\userComment{Shai}{#1}}
\newcommand{\ld}[1]{\userComment{Lena}{#1}}
\newcommand{\ah}[1]{\userComment{Alon}{#1}}
\newcommand{\ns}[1]{\userComment{Noam}{#1}}
\newcommand{\yk}[1]{\userComment{Yoav}{#1}}
\newcommand{\cpesystem}{CPE\xspace}

\newcommand{\squishlist}{
 \begin{list}{$\bullet$}
  { \setlength{\itemsep}{0pt}
     \setlength{\parsep}{1pt}
     \setlength{\topsep}{1pt}
     \setlength{\partopsep}{0pt}
     \setlength{\leftmargin}{0.2em}
     \setlength{\labelwidth}{0.1em}
     \setlength{\labelsep}{0.1em} 
    } 
}
 \newcommand{\squishend}{\end{list}}
\newcommand{\eat}[1]{}

\begin{abstract}
Prompts are how humans communicate with LLMs. Informative prompts are essential for guiding LLMs to produce the desired output. However, prompt engineering is often tedious and time-consuming, requiring significant expertise, limiting its widespread use. We propose Conversational Prompt Engineering (CPE), a user-friendly tool that helps users create personalized prompts for their specific tasks. CPE uses a chat model to briefly interact with users, helping them articulate their output preferences and integrating these into the prompt. The process includes two main stages: first, the model uses user-provided unlabeled data to generate data-driven questions and utilize user responses to shape the initial instruction. Then, the model shares the outputs generated by the instruction and uses user feedback to further refine the instruction and the outputs. The final result is a few-shot prompt, where the outputs approved by the user serve as few-shot examples. A user study on summarization tasks demonstrates the value of CPE in creating personalized, high-performing prompts. The results suggest that the zero-shot prompt obtained is comparable to its -- much longer -- few-shot counterpart, indicating significant savings in scenarios involving repetitive tasks with large text volumes.
\end{abstract}

\section{Introduction}
Large Language Models (LLMs) offer significant advantages for enterprises by enabling a wide range of applications, such as automating customer support, 
Retrieval Augmented Generation (RAG), 
text summarization, and more. 
To maximize the capabilities of LLMs, crafting effective prompts is essential.
The prompts should clearly define the 
task and desired outcomes, conveying 
nuances, preferences, and important output requirements. 
Such tailored prompts help the LLM understand the precise nature of the task, guiding it to generate more satisfactory results. 
However, the process of designing and optimizing prompts, known as prompt engineering (PE), is computationally demanding and time-consuming.
It requires a deep 
understanding of how LLMs interpret and respond to 
instructions, 
as well as 
to anticipate how slight changes in phrasing or context might affect the output, making it a complex task even for experienced practitioners.

 
To reduce this burden, recent works explore automating PE 
\cite{zhou2022large,pryzant-etal-2023-automatic}.
However, 
they 
typically assume 
access to labeled data, which is often 
challenging to obtain, especially for generation tasks. 
Moreover, these methods often 
require users to supply manually-created seed 
prompts, which is also far from trivial. 

Here, we introduce an alternative approach -- Conversational Prompt Engineering (\cpesystem) -- that removes the need for labeled data and initial prompts. The key idea is to utilize chat models to assist users in creating prompts through a brief and user-friendly conversation, considering a small set of unlabeled data provided by the user.

\cpesystem is based on three core insights. First, users frequently struggle to clearly define their task and their precise expectations from the model responses. Advanced chat models can assist users in better understanding and articulating their exact requirements, making it easier for users to communicate their needs. Second, given a task, unlabeled input texts can be leveraged by LLMs to suggest 
data-specific dimensions of potential output preferences. This process can 
further help 
users specify their task requirements by highlighting relevant aspects of the data. Finally, user feedback on specific model-generated outputs can be leveraged not only to improve the outputs themselves, but also to refine the instruction to be applied to unseen texts.

The output prompt generated by \cpesystem is a few-shot (FS) prompt, where the preferences and intentions expressed by the user throughout the chat are incorporated into the instruction, and the outputs approved by the user serve as the FS examples.
 
\cpesystem is specifically designed for users who need to perform the same task repeatedly on a large volume of text elements. This is often the case in enterprise scenarios, such as summarizing email threads or writing personalized advertising content.
Our \cpesystem implementation can be used for any task defined by the user, such as topic generation, advertising content generation, creative writing etc.  However, due to the limited number of annotators, in the evaluation section we focus on the common task of summarization. 

The rest of the paper is organized as follows: We present an overview of \cpesystem from the perspective of a user in Section~\ref{sec:user_workflow}; We describe the design and implementation of \cpesystem in Section~\ref{sec:design}; and finally we 
share the results of 
a user-study demonstrating the effectiveness of \cpesystem in Section~\ref{sec:evaluation}.

\section{Related Work}

Given the significant role 
of prompts in LLMs performance, 
there has been growing interest in developing strategies and algorithms to enhance their quality. One such strategy is Chain-of-Thought prompting, first proposed by \citet{wei2022chain}. This approach involves enriching in-context examples with reasoning steps to improve performance on complex tasks.

Another line of studies 
focus on automatic prompt improvement. In \citet{shin2020autoprompt}, 
trigger-tokens 
are used to automatically optimize prompts in Masked Language Models. Another approach is an iterative process in which an LLM initially generates prompts (e.g., based on seed examples), which are then automatically evaluated and sent back to the LLM for further rephrasing \cite{zhou2022large, yang2023large}. \citet{ye2023prompt} propose constructing a meta-prompt to guide the LLM during the automatic PE process.
\citet{xu2023expertprompting} present expert prompting, where automatically-generated expert identities are added to the instruction given to the LLM to produce an expert answer. \citet{fernando2023promptbreeder} introduce Promptbreeder, a self-referential, self-improving mechanism for enhancing prompts. Promptbreeder repeatedly mutates prompts, evaluates their effectiveness, and generates improved versions of both the task prompts and the mutation prompts.

Other methods 
involve human-in-the-loop mechanisms. \citet{evallm2024} introduce EvalLM, an interactive system designed for iteratively refining prompts by evaluating multiple outputs based on user-defined criteria. EvalLM not only aids in enhancing the prompts themselves but also improves the evaluation criteria used to assess these prompts. While the motivation of this work bears similarity to ours, \cpesystem differs by providing a more natural chat-based interface, suggesting preference criteria and discussing them with the user; and crucially leverages LLM capabilities to provide the prompt revisions output enhancements explicitly, which the user can then approve.

In parallel to these studies, 
many tools have been created to assist users in leveraging LLMs like ChatGPT across different scenarios. 
Most relevant to our work are frameworks 
focused on developing and sharing prompts for various use cases.
One example is PromptBase, a marketplace where users can purchase GPT prompts for various tasks. The site also offers an interactive chat tool for creating prompts, similar to our proposal.\footnote{\url{https://promptbase.com/prompt/interactive-prompt-generator-2}} However, PromptBase operates in zero-shot (ZS) mode, relying on a single initial prompt. It further 
requires the user to have the expertise to guide the conversation and provide information for prompt creation. In contrast, \cpesystem uses multiple 
pre-defined prompts to direct the model through a 
well-structured flow, reducing the user's cognitive burden and effectively gathering relevant user's feedback.

\section{System Overview}
\label{sec:user_workflow}
 
\subsection{Terminology}


\textbf{\cpesystem}: The \cpesystem implementation described in this paper, which includes a user interface (UI) and a three-party chat backbone, both of which are detailed further.\\
\textbf{Target model}: The model for which the prompt is formatted.\\
\textbf{Instruction}: A description of the task and required output. Note, the instruction does not include the 
FS examples. \\ 
\textbf{CPE instruction}: The final instruction proposed by \cpesystem and approved by the user.\\
\textbf{Prompt}: The input given to the LLM, which includes the instruction formatted according to the target model template.\\
\textbf{\cpesystem ZS prompt}: The prompt that includes the final instruction proposed by \cpesystem and approved by the user, excluding input-output examples.\\
\textbf{\cpesystem FS prompt}: Similar to the \cpesystem ZS prompt, but includes input-output examples that have been approved by the user.\\
\textbf{User data}: A set of at least three unlabeled input examples,  
representing the user's data, which is provided to \cpesystem by the user.\\ 
\textbf{Prompt Outputs}: The results produced by the target model when using a prompt generated by \cpesystem applied to the user data. 

\subsection{Overview}

The ultimate goal of \cpesystem is to help the user create a prompt that, when fed to the user's model, will generate outputs aligned with the user's requirements.
Next, 
we provide an overview of \cpesystem from the user's perspective. A chat with \cpesystem typically consists of several key stages:
\begin{enumerate}
    \item \textbf{Initialization}: 
    The initialization step is composed of two parts: \textbf{target model selection}, where the user selects the target model; and \textbf{user data initialization}, where an unlabeled data file is uploaded by the user.
    \item \textbf{Initial Discussion and First Instruction Creation}: \cpesystem analyzes three examples from the user data,\footnote{We set the number of input examples to three due to the limited context length.} engages with the user in a discussion about various data-aware aspects of their task and output preferences, and generates an initial instruction. For example, assuming a user uploaded three movie reviews and would like to summarize them, \cpesystem -- 
    after analyzing the user data -- 
    may ask the user whether he/she would like to focus the summaries on the movie's plot or the reviewer's opinion. If the user indicates interest in both, \cpesystem can then suggest the following 
    initial instruction: 
    \emph{Provide a brief summary that includes the plot and the reviewer’s opinion.}
    \item \textbf{Instruction Refinement}: 
    The instruction is revised by \cpesystem based on user feedback, which can be provided either w.r.t the proposed instruction or the prompt outputs. E.g, assuming the user has further requested 
    summaries in bullet points, \cpesystem may propose the following instruction: \emph{Provide a brief summary that includes the plot and the reviewer’s opinion in bullet points.}
    \item \textbf{Output Generation}: The latest instruction is used to construct a prompt, which is then provided to the target model to generate the corresponding prompt outputs.
    \item \textbf{User Feedback and Output Enhancement}: The user provides feedback on the prompt outputs. If the user approves them as they are, no further refinements are needed, and the system proceeds to the final stage. If not, \cpesystem improves the outputs based on the feedback until they are accepted by the user, then suggests a modified instruction and returns to the instruction refinement stage.      
    \item \textbf{Convergence on CPE FS Prompt}: A chat ends after the instruction and all prompt outputs are approved by the user. Subsequently, the CPE FS prompt is shared.
\end{enumerate}

\begin{figure}[t]
\begin{center}
\includegraphics[width=0.85\columnwidth]{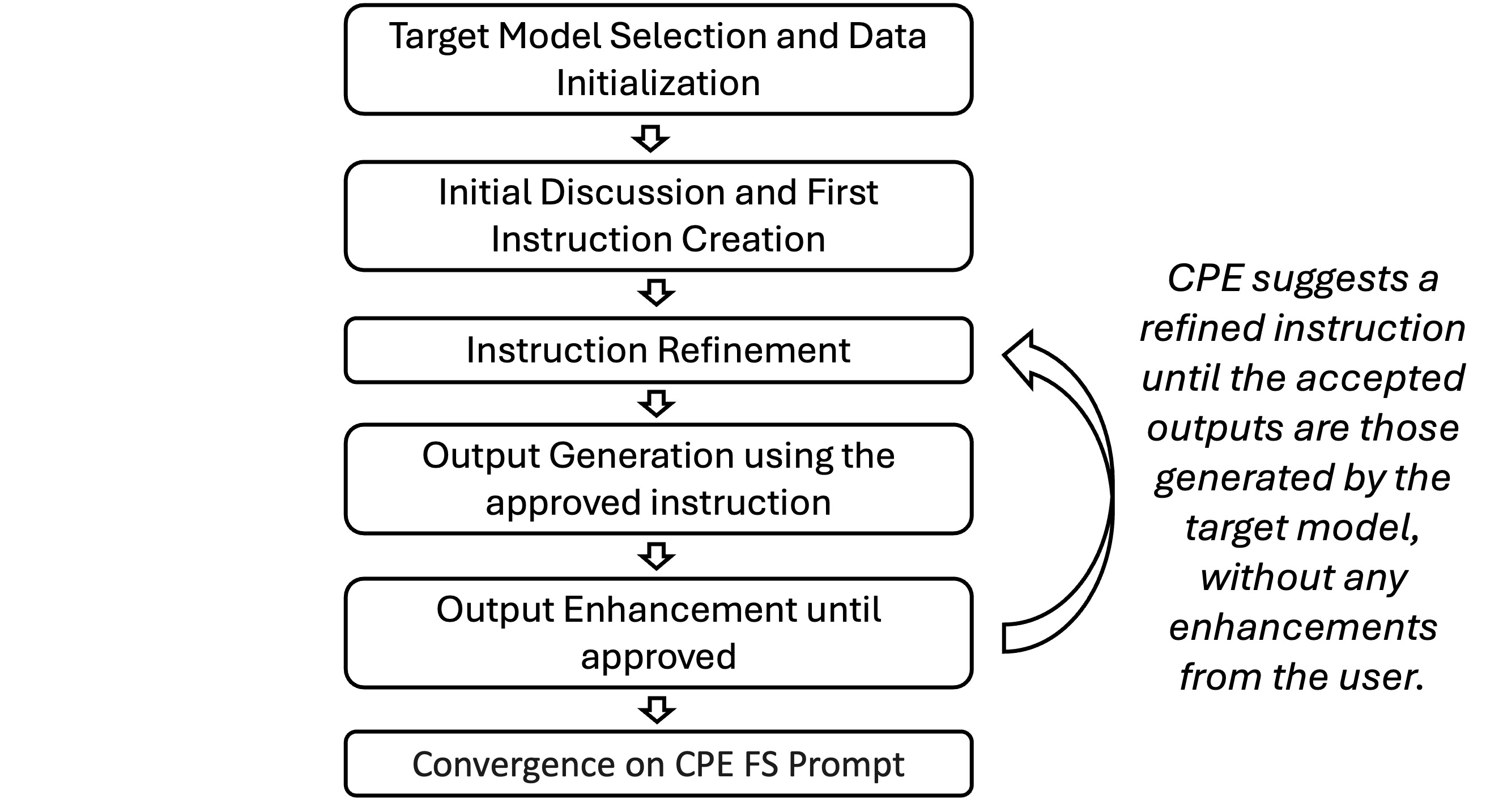}
\caption{CPE Workflow from the user's perspective. 
Each step can be a multi-turn conversation between the user and \cpesystem.}
\label{fig:user_workflow}
\end{center}
\end{figure}

The flow described above is depicted in Figure~\ref{fig:user_workflow}.  Note that tages $3$-$5$ can be repeated iteratively before reaching stage $6$. The aim of the iterations is to refine the instruction until the \textit{prompt outputs} (as opposed to the enhanced versions generated by \cpesystem) are accepted by the user. Moreover, while the order of stages detailed above represents a typical chat from start to finish, \cpesystem maintains complete flexibility, resembling any chat that one would produce with a SoTA LLM. Thus, 
at any stage a user may jump between input examples, ask questions about the prompt, or end the chat arbitrarily.

\subsection{UI}

\cpesystem is accessed via a web interface which contains three modules: chat, survey, 
and evaluation. Chat is the core module where the interaction with \cpesystem occurs. The UI allows the user to perform actions beyond the actual chat, e.g., upload user data, or download the CPE FS prompt.

\section{Design and Implementation}
\label{sec:design}


\subsection{Three-party Chat}
\label{sec:three_party_chat}

\begin{figure}[t]
\begin{center}
\includegraphics[width=1\columnwidth]{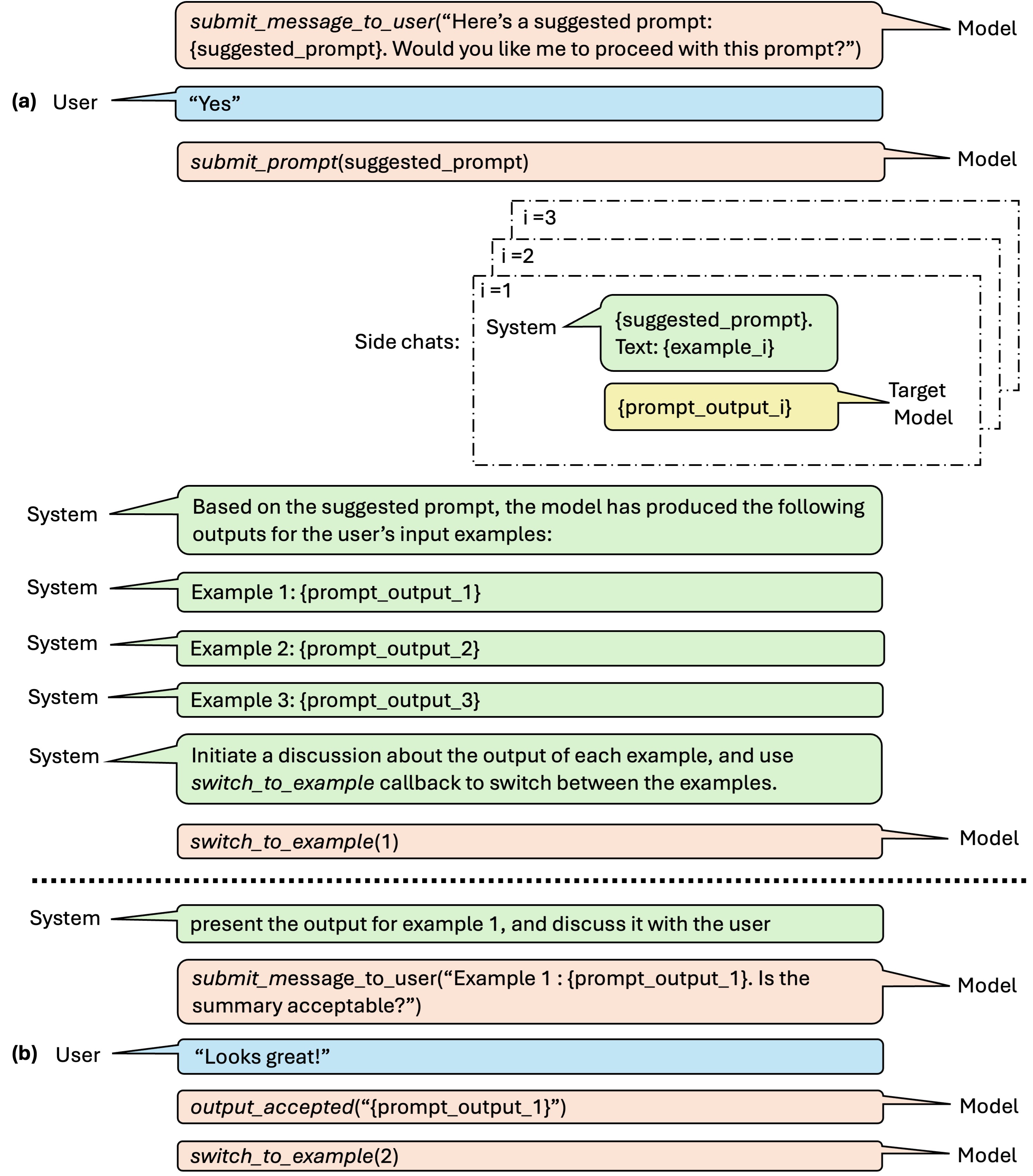}
\caption{Messages exchanged between the different actors in a chat with \cpesystem, starting from (a) the user approves a suggested prompt, until (b) user approves the output of the first example. System messages are abbreviated (see full text in Appendix~\ref{sec:model_instruction}). We use \textit{suggested\_prompt}, \textit{example\_i} and \textit{prompt\_output\_i} to denote the prompt suggested by \cpesystem, the $i^{th}$ example, and its prompt output respectively. All the messages below the dotted line are sent with a filtered context. }

\label{fig:in_depth}
\end{center}
\end{figure}


The interaction within \cpesystem embodies three actors: the user, the system, and the model. 
    \paragraph{The user:} communicates with \cpesystem 
    via
    the UI. 
    \paragraph{The model:} The LLM that handles the core capabilities of \cpesystem: user data analysis and discussion, instruction refinement, and output enhancement. The LLM is instructed to respond in a particular format of one of several \textit{API calls}. Each response by the model is then executed, and may or may not involve sharing a message with the user (see Section~\ref{callbacks}). Out implementation uses Llama-3-70B. 
    \paragraph{The system:} 
    orchestrates the interaction between the user and the model. Each model response (API call), triggers an action by the system. Contrary to common practice -- where the model is assigned a single initial system instruction -- we invoke multiple system instructions, {\it dynamically\/} 
    throughout the chat. This mechanism allows us greater flexibility and control over the model, and enables the 
    side-chats (see below). 

\subsection{Context Management} 
We construct the context of each message from the system to the model individually. 
We implement a context filtering mechanism that enables us to control the context provided to the model at each stage of the conversation.
One of the motivations for this mechanism is the context length limitations. As detailed in Section~\ref{sec:user_workflow}, after generating the initial instruction, the model engages -- via the system -- 
with the user to discuss the outputs of three text examples. While discussing each example, we use a filtered context that does not 
contain messages that are exchanged when 
discussing 
other examples. 

Another use of the mechanism is for 
\textbf{Side-chats}.
In various stages of the \cpesystem session, it is necessary to invoke a side-chat that includes only specific parts from the full chat. A prominent example is when \cpesystem aims to mimic real-world usage of the considered \cpesystem prompts with the target model, that naturally can not include the context of the conversation used to develop the prompt. Thus, to properly test the effectiveness of a considered prompt outside of \cpesystem, the system uses a side-chat with the target model, where the examined prompt is the only context.

\subsection{Chain of Thought (CoT) \label{COT}} 
We use CoT \cite{wei2022chain} in side-chats during various stages where more careful guidance of the model is needed. For example, after the user provides feedback for prompt outputs, the model needs to understand if and how to refine the instruction accordingly. Due to the complexity of this stage, we use CoT by first asking the model to summarize the comments made by the user in a side-chat (showing only those comments as context); once these summaries are available, we ask the model to use them to suggest a new instruction. 

\subsection{API Calls} \label{callbacks}

As mentioned in Section~\ref{sec:three_party_chat}, the model is guided to respond at each turn using pre-defined API calls. Next, we describe these callback functions.\\
\paragraph{submit\_message\_to\_user(msg):}
Used for any message that is meant to be shared with the user.\\ 
\paragraph{submit\_prompt(prompt):} Called after the user approves \emph{instruction} proposed by the model.     
The system then takes the following actions: 1) Use a context-free side-chat to prompt the \emph{target model} to generate outputs for the user data.
2) Share prompt outputs with the model, and instruct it to discuss them with the user (see Figure~\ref{fig:in_depth}).\\
\paragraph{switch\_to\_example(example\_num):} Called when moving to discuss the prompt output of example <example\_num>. 
The system instructs the model to discuss it with the user.\\ 
\paragraph{output\_accepted(example\_num,output):} Called after the user approves 
the latest output of example <example\_num>. 
In case there are remaining outputs to discuss, \emph{switch\_to\_example(example\_num$+1$)} is triggered
by the system. 
Otherwise, \emph{end\_outputs\_discussion()} is triggered.\\
\paragraph{end\_outputs\_discussion():} 
Called after all prompt outputs are approved by the user. 
Correspondingly, the system uses 
CoT to analyze feedback given by the user on the original outputs, aiming to modify the instruction if needed. A new instruction is generated in a side-chat with a partial context that includes only the discussion between the user and the model about the outputs.  
This revised instruction is then shared with the model, which either shares it with the user,
or suggests to end 
the conversation (via \emph{conversation\_end()}).\\
\paragraph{conversation\_end():} Called after the user approved the prompt and all prompt outputs. 
The system then instructs the model to end the conversation and share the CPE FS prompt.

\subsection{\cpesystem as an Agentic Workflow}
Agentic workflows is an important emerging trend, enabling LLMs to autonomously perform complex tasks. \cpesystem can be viewed as an LLM-based agent with human-in-the-loop (the user). Next, we outline its key components, and their realization within \cpesystem:

\paragraph{Planning:} \cpesystem's planning is done by decomposing the full task of generating a personalized prompt into multiple sub-tasks, as illustrated  
in Figure \ref{fig:in_depth}. Reflection and CoT techniques are also employed, as discussed in Section \ref{COT}. From the implementation perspective, the prompts injected by the system actor throughout the conversation, communicate the sub-tasks to the 
LLM. 

 \paragraph{Tools:} \cpesystem utilizes several tools via the API calls. These include the prompt execution tool, which is called by the submit\_prompt API. 
Part of the system actor instructions include placeholders for information obtained from the tools. This mechanism enables the 
model 
to utilize this information within the chat. For example, the outputs generated by the target model are included in the system message 
to the model, 
as shown in Figure \ref{fig:in_depth}.

\paragraph{Memory:} The context management component of \cpesystem ensures effective performance within the memory limitations of the LLM context length.


\section{User Study}
\label{sec:evaluation}
To assess the benefit of \cpesystem in practical scenarios we conducted a user study, focused on summarization tasks, as these tasks encompass a broad spectrum of potential user preferences.

\textbf{Method.} We engaged $12$ participants to interact with \cpesystem and develop prompts for a summarization use-case. Each participant, well-versed in 
PE, 
selected data from a catalog of unlabeled datasets, listed in Section~\ref{sec:user_study}. Participants were encouraged to engage in open-ended conversations with \cpesystem, presenting any summarization requirements they desired. Following their interactions with \cpesystem, 
they completed a brief 
survey, rating 
their agreement with four statements on a Likert scale 
ranging from 1 (strongly disagree) to 5 (strongly agree). The assessed dimensions included satisfaction with the 
obtained 
\cpesystem instruction, 
benefits from the joint conversation, 
overall pleasantness of the chat, and convergence time.

Participants were then asked to evaluate the final \cpesystem prompt using an unlabeled test split from the respective dataset.\footnote{Note that despite the availability of a labeled test split for some datasets, utilizing it for automatic evaluation using standard metrics is less relevant here, given that our user study focuses on individual preferences.} 
For each test example, $3$ 
summaries were generated using the following prompts: 1) a baseline prompt (\emph{baseline}) (see in \ref{sec:user_study} how it was generated); 2) the CPE ZS prompt; and 3) the CPE FS prompt. Participants were asked to select the best and worst summaries per example. 
The prompts used to generate each summary were concealed from the participants, and the summaries were presented in a random order.
For more details, see Section~\ref{sec:user_study} in the Appendix.

\textbf{Results.} The survey results are presented in Table~\ref{tab:survey_results}, confirming the benefits of using \cpesystem to create prompts that effectively meet user requirements. 
Note 
that participants rated convergence time slightly lower as it took $25$ minutes on average to converge on the CPE FS prompt. This may be partly attributed to the complexity of the summarization criteria provided by participants. We aim to address this issue and improve convergence time in future work.

\begin{table}[!ht]
    \small
    \centering
    \begin{tabular}{l|l}
    \hline
        Satisfaction from CPE instruction & 4.6 \\ \hline
        Benefit from joint conversation & 4.5 \\ \hline
        Chat pleasantness & 4.8 \\ \hline
        Convergence time & 3.8 \\ \hline
    \end{tabular}
    \caption{Survey results.}
\label{tab:survey_results}
\end{table}

\begin{figure}[H]
\begin{center}
\includegraphics[width=0.97\columnwidth]{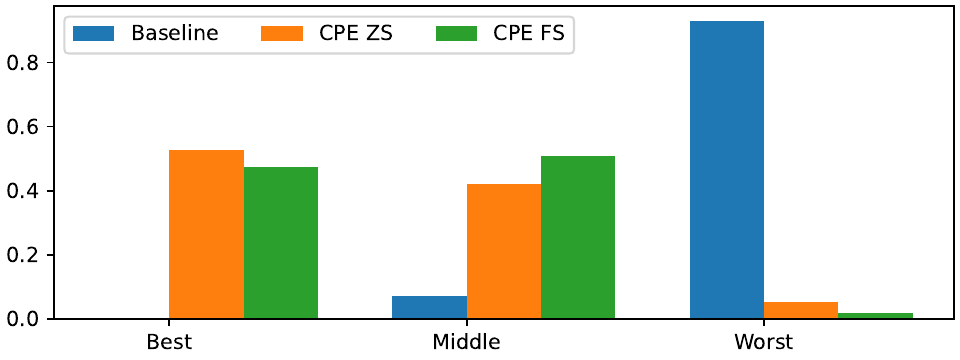}
\caption{Evaluation results. The frequency of the three generated summaries across the three ranking categories: Best/Middle/Worst.}
\label{fig:manual_chat_summary_selection}
\end{center}
\end{figure}

Figure~\ref{fig:manual_chat_summary_selection} presents the evaluation results, showing the distribution of the three summaries for each rank (best, middle, and worst) as chosen by users across individual test examples. These results are aggregated from $12$ chat sessions of all participants, with a total of $112$ examples.
%
%
Overall, participants preferred summaries generated by the \cpesystem prompts over those produced by the \emph{baseline} prompt. When comparing user preferences between the CPE ZS and CPE FS prompts, their summaries were ranked as the best in 53\% and 47\% of instances, respectively.\footnote{This difference was found to be statistically insignificant, as determined by a one-sample T-test \cite{drummond2011statistics}.}
The comparable satisfaction of participants with these prompts suggests that \cpesystem effectively integrated user preferences into the prompt, rendering the few-shot examples unnecessary. This capability of \cpesystem can save users substantial costs when performing repetitive tasks on large volumes of text, as it significantly reduces the number of tokens in the prompt.

\textbf{Analysis.} 
We further analyze the interactions users had with \cpesystem. 
On average, these chats required $32 (\pm9.6)$ turns to reach convergence. In $4/12$ chats, the final instruction was the one created by at the end of the initialization step, based on the analysis of the unlabeled data. In the remaining chats, instructions were refined through multiple iterations, indicating that two-thirds of users utilized 
the output enhancement phase 
to improve their instruction.
 
Additionally, we compare the initial and final \cpesystem instructions. We find that for $7/12$ chats, the distance between these two instructions is $\ge 50$ characters, with an average distance of $96.3 (\pm108.3)$ across all $12$ chats, indicating substantial differences that emerged throughout the chat (see Table~\ref{tab:analysis_example} for an example). This analysis underscores the potential value of using \cpesystem to refine instructions, achieving a detailed and nuanced resolution of task requirements.

\begin{table}[!ht]
    \small
    \centering
    \begin{tabular}{p{0.5cm}|p{5.5cm}}
    \hline
        Turn & Instruction \\ \hline
        $6$ & \emph{Generate a list of main claims for each debate, grouped by topic, in 1-2 concise sentences per claim.}\\ \hline
        $24$ & \emph{Generate a list of main claims for each debate, grouped by topic, in 1-2 concise sentences per claim, \textbf{without specific examples, and break down complex or multi-topic claims into simpler separate ones.}}\\ \hline
        $34$ & \emph{Generate a list of main claims for each debate, grouped by topic, in 1-2 concise sentences per claim, without specific examples, breaking down complex or multi-topic claims into simpler separate ones, \textbf{and avoiding redundant or repetitive claims.}} \\ \hline
    \end{tabular}
    \caption{An example of how an instruction was refined throughout a chat with \cpesystem.}
\label{tab:analysis_example}
\end{table}

\section{Discussion}
We introduced \cpesystem, a novel approach to help users with the critical task of PE, 
particularly those who are interested in repetitive tasks across large datasets. 
Our user study, centered on the complex task of summarization, demonstrated \cpesystem's effectiveness in generating efficient prompts that produce responses that align 
with users' expectations.

An interesting open question is whether  the prompts created by \cpesystem can be further improved by using them as initial prompts for existing automatic 
PE 
methods, such as \cite{ye2023prompt}.
In addition, it would be intriguing to extend the application of \cpesystem beyond prompt generation. In light of the recent surge in interest around LLM-empowered agent systems, it would be valuable to explore whether the \cpesystem approach could assist users in planning and creating agentic workflows.



\bibliography{custom, anthology}

\appendix

\section{User Study}
\label{sec:user_study}

\textbf{Setup.} The participants of the user study included $11$ researchers that were asked to select a dataset from a closed catalog we uploaded to the system. Each such dataset had $2$ splits, one training -- $3$ examples used during the chat -- and one for evaluation, where up to $8$ examples were used.

For the dataset catalog we used samples from the following datasets: the Space class of 20 Newsgroups \cite{Lang1995NewsWeederLT}, TL;DR \cite{tldr}, debate speeches \cite{orbach-etal-2020-echo}, and two collections from Wikipedia that we curated (articles under the movies and animals categories).

As a baseline generic prompt we used the following instruction: \emph{Summarize the main points and key information from the provided text in a concise and clear manner, preserving the original meaning and content.} This prompt was obtained by prompting Llama-3-70B-instruct with following text: \emph{Generate a concise and general prompt for a summarization task in one sentence}.

\textbf{Survey.} We used the following statements in the survey:

\textbf{Satisfaction from CPE instruction}: \emph{I'm satisfied with the final prompt <CPE\_prompt>, it met my requirements.}

\textbf{Gain from thinking process}: \emph{The system helped me think through how the desired outputs should look like and what criteria to consider when building the prompt.}

\textbf{Chat pleasantness}: \emph{I felt the system was pleasant and responsive throughout the interaction.}

\textbf{Convergence time}: \emph{I'm satisfied with the time it took to come up with the final prompt.}

\section{System Instructions}
\label{sec:model_instruction}
This is the full list of instructions that are provided by the system to the model in different parts of the chat.

\textbf{Initialization}: {\small You and I (system) will work together to build a prompt for the task of the user via a chat with the user. This prompt will be fed to a model dedicated to perform the user's task. Our aim is to build a prompt that when fed to the model, produce outputs that are aligned with the user's expectations. Thus, the prompt should reflect the specific requirements and preferences of the user from the output as expressed in the chat. You will interact with the user to gather information regarding their preferences and needs. I will send the prompts you suggest to the dedicated model to generate outputs, and pass them back to you, so that you could discuss them with the user and get feedback. User time is valuable, keep the conversation pragmatic. Make the obvious decisions by your Don't greet the user at your first interaction.You should communicate with the user and system ONLY via python API described below, and not via direct messages. The input parameters to API functions should be string literals using double quotes. Remember to escape double-quote characters inside the parameter values. Note that the user is not aware of the API, so don't not tell the user which API you are going to call. Format ALL your answers python code calling one of the following functions:}
        
\textbf{API instruction}: {\small function submit$\_$message$\_$to$\_$user(msg): call this function to submit your message to the user. Use markdown to mark the prompts and the outputs. 
        
function submit$\_$prompt(prompt): call this function to inform the system that you have a new suggestion for the prompt. Use it only with the prompts approved by the user.

function switch$\_$to$\_$example(example$\_$num): call this function before you start discussing with the user an output of a specific example, and pass the example number as parameter.

function show$\_$original\_text(example$\_$num): call this function when the user asks to show the original text of an example, and pass the example number as parameter.

function output$\_$accepted(example$\_$num, output): call this function every time the user unequivocally accepts an output. Pass the example number and the output text as parameters.

function end$\_$outputs$\_$discussion(): call this function after all the outputs have been discussed with the user and all 3 outputs were accepted by the user.

function conversation$\_$end(): call this function when the user wants to end the conversation.}

\textbf{User data intro}: {\small The user has provided some text examples. I've selected a few of them that you will use in the conversation. Note that your goal to build a generic prompt, and not for these specific examples.}

\textbf{User data analysis}: {\small Before suggesting the prompt, briefly discuss the text examples with the user and ask them relevant questions regarding their output requirements and preferences. Please take into account the specific characteristics of the data. Your suggested prompt should reflect the user's expectations from the task output as expressed during the chat. Share the suggested prompt with the user before submitting it. Remember to communicate only via API calls.}

\textbf{Outputs intro}: {\small Based on the suggested prompt, the model has produced the following outputs for the user input examples:}

\textbf{Outputs analysis}: {\small For each of 3 examples show the model output to the user and discuss it with them, one example at a time. Use switch\_example API to navigate between examples. The discussion should take as long as necessary and result in an output accepted by the user in clear way, with no doubts, conditions or modifications. When the output is accepted, call output\_accepted API passing the example number and the output text. After calling output\_accepted call either switch\_to\_example API to move to the next example, or end\_outputs\_discussion API if all 3 have been accepted. Assume that the user comments relay to the output. Only when the user explicitly says that he wants to update the prompt and not the output, show the updated prompt to them. Remember to communicate only via API calls.}

\textbf{Example switch}: {\small You have switched to <i>. Look at the user comments and the accepted outputs for the previous examples, apply them to the model output of this example, and present the result to the user. Indicate the example (number), and format the text so that the output and your text are separated by empty lines. Discuss the presented output taking into account the system conclusion for this example if exists.}

\textbf{Outputs CoT start}: {\small In the following discussion, the user was asked to give feedback on the model's outputs that were generated by the prompt "PROMPT". The outputs that did not meet the user's requirements were modified.}

\textbf{Outputs CoT end}: {\small Analyze the conversation above, and share the comments made by the user on Examples 1-3. Any comment should be shared, even if minor. If no comment has been made, accept the prompt. If any comment has been made, recommend how to improve the prompt so it would produce the accepted outputs directly.}
    
\textbf{Outputs discussion}: {\small Continue your conversation with the user. Do the recommendations above suggest improvements to the prompt? If so, present the modified prompt to the user, and submit it only after the user approve it. Otherwise, if no modifications to the prompt are required, communicate it to user and suggest to finish the conversation.}

\textbf{Conversation end}: {\small This is the end of conversation. Say goodbye to the user, and inform that the final prompt that includes few-shot examples and is formatted for the <model> can be downloaded via **Download few shot prompt** button below. Also, kindly refer the user to the survey tab that is now available, and let the user know that we will appreciate any feedback.}

\end{document}